\documentclass[conference]{IEEEtran}
\IEEEoverridecommandlockouts
\usepackage{cite}
\usepackage{amsmath,amssymb,amsfonts}
\usepackage{booktabs, multirow}
\usepackage{algorithmic}
\usepackage{graphicx}
\usepackage{textcomp}
\usepackage{threeparttable}
\usepackage{xcolor}
\usepackage{url}
\usepackage{tablefootnote}
\usepackage{subcaption}

\def\BibTeX{{\rm B\kern-.05em{\sc i\kern-.025em b}\kern-.08em
    T\kern-.1667em\lower.7ex\hbox{E}\kern-.125emX}}

\begin{document}

\title{Quantifying Individual and Joint Module Impact in Modular Optimization Frameworks\\
\author{\IEEEauthorblockN{Ana Nikolikj, Ana Kostovska, Diederick Vermetten, Carola Doerr and Tome Eftimov}
\thanks{Ana Nikolikj (Email: ana.nikolikj@ijs.si) and Tome Eftimov (Email: tome.eftimov@ijs.si) are with Computer Systems Department, Jo\v{z}ef Stefan Institute, Ljubljana, Slovenia.}
\thanks{Ana Kostovska (Email:ana.kostovska@ijs.si) is with the Department of Knowledge Technologies, Jo\v{z}ef Stefan Institute, Ljubljana, Slovenia.}
\thanks{Ana Nikolikj. and Ana Kostovska are also with the Jo\v{z}ef Stefan International Postgraduate School, Ljubljana, Slovenia.}
\thanks{Diederick Vermetten (Email:
d.l.vermetten@liacs.leidenuniv.nl) is with the Leiden University, LIAC, Leiden, Netherlands.}
\thanks{Carola Doerr (Email: carola.doerr@lip6.fr) is with the Sorbonne Université, CNRS, LIP6, 1000 Paris, France.}
\thanks{The authors acknowledge the support of the Slovenian Research Agency through program grants P2-0098 and P2-0103, young researcher grants No. PR-12897 to AN and  PR-09773 to AK, project No. J2-4460, and a bilateral project between Slovenia and France grant No. BI-FR/23-24-PROTEUS-001 (PR-12040).}
}
}

\maketitle

\begin{abstract}
This study explores the influence of modules on the performance of modular optimization frameworks for continuous single-objective black-box optimization. There is an extensive variety of modules to choose from when designing algorithm variants, however, there is a rather limited understanding of how each module individually influences the algorithm performance and how the modules interact with each other when combined. We use the functional ANOVA (f-ANOVA) framework to quantify the influence of individual modules and module combinations for two algorithms, the modular Covariance Matrix Adaptation (modCMA) and the modular Differential Evolution (modDE). We analyze the performance data from 324 modCMA and 576 modDE variants on the BBOB benchmark collection, for two problem dimensions, and three computational budgets. Noteworthy findings include the identification of important modules that strongly influence the performance of modCMA, such as the~\textit{weights\ option} and~\textit{mirrored} modules for low dimensional problems, and the~\textit{base\ sampler} for high dimensional problems. The large individual influence of the~\textit{lpsr} module makes it very important for the performance of modDE, regardless of the problem dimensionality and the computational budget. When comparing modCMA and modDE, modDE undergoes a shift from individual modules being more influential, to module combinations being more influential, while modCMA follows the opposite pattern, with an increase in problem dimensionality and computational budget.

\end{abstract}


\section{Introduction}
\label{sec:intoduction}
A core application of evolutionary algorithms is the optimization of continuous single-objective black-box problems. A plethora of algorithms already exist~\cite{EibenS03,StorkEB22taxonomy}, and as the field matures researchers are increasingly focusing on refining existing algorithms rather than developing new algorithmic paradigms~\cite{dreo2021paradiseo, de2021tuning, DBLP:journals/corr/abs-2304-09524, camacho2021pso}. For example, in the data from the BBOB workshops~\cite{BBOBdata} we can find more than 50 variants of the CMA-ES algorithm~\cite{hansen2001completely}. 

Rather than studying algorithm variants in isolation, it was proposed in~\cite{de2021tuning, DBLP:journals/corr/abs-2304-09524} to consider a standardized~\textit{modular optimization framework} where various algorithm variants can be systematically assessed, ensuring uniformity in implementation details across all variants. The main idea behind modular optimization frameworks is to break down an algorithm into smaller components called~\textit{modules}, which can then be configured by taking on different options that influence the algorithm's behavior. The modules can be seamlessly integrated to form new algorithm variants. Additionally, each module operates independently, so it can be removed without affecting the functionality of the rest. The above-mentioned studies emphasize the advantages of modular algorithms, both in terms of fair comparisons from an implementation perspective and also the possibility of examining the interactions between different modules. 

The challenge that remains is how to quantify the influence of individual modules and the influence of module interactions on the algorithm performance. Often, algorithm designers manually assess module influence by investigating the local neighborhood of a given algorithm variant, for example, a common approach is to alter one module at a time and to then observe the changes in performance. Probably even more common is to simply compare the performance of an algorithm variant to a default implementation of the same algorithm, or compare it to a limited set of other variants. The only information obtained with this analysis is how the different module options perform in the context of a few other variants, analyzing only the individual influence of a module and ignoring the interaction influence that results from the combination with the other modules.

\textbf{Our contribution:} In this study, we employ the f-ANOVA framework proposed in~\cite{hutter2014efficient} to quantify individual and interaction module effects in modular optimization frameworks. Our study involves the performance data of the modCMA~\cite{de2021tuning} and modDE~\cite{DBLP:journals/corr/abs-2304-09524} frameworks, evaluated on the 24 problems from the BBOB benchmark suite, in dimensions 5 and 30, for three different computational budgets. By analyzing 324 variants of modCMA and 576 variants of modDE, we assess individual and interaction module effects through two scenarios: one evaluating the module's effects on the level of an entire benchmark suite and another examining their effects on a problem-specific level. The analysis reveals important modules and combinations of modules, such as~\textit{weights option} and~\textit{mirrorred} in modCMA for dimension 5 and~\textit{base sampler} for dimension 30. Additionally, it highlights the importance of individual modules such as~\textit{lpsr}, over module combinations in modDE, particularly for lower budgets. Overall, the findings provide valuable insights for algorithm designers and suggest potential applications for problems where module combinations contribute to algorithm performance.

\textbf{Outline:} Section~\ref{sec:related_work} reviews relevant research on module importance in modCMA and modDE, together with studies from related domains that utilize f-ANOVA to estimate the importance of hyperparameters. Section~\ref{sec:fANOVA} outlines the functional ANOVA methodology. The experimental design is explained in Section~\ref{sec:experiments}, followed by a discussion of key results in Section~\ref{sec:results}. Section~\ref{sec:conclusion} concludes the paper.

\textbf{Data and code availability:} The data and the code involved in this study are available at~\cite{GitRepository}.

\section{Related work}
\label{sec:related_work}
Initial work in this direction~\cite{van2017algorithm} involves conducting a comprehensive assessment of the importance of modules in CMA-ES. This was achieved by utilizing data from a complete enumeration of all module options to analyze the performance contribution of each module. However, this method proves unfeasible when dealing with an increasing set of modules.

Both~\cite{de2021tuning} and~\cite{DBLP:journals/corr/abs-2304-09524} utilize an algorithm tuning tool called~\textit{irace}~\cite{lopez2016irace} to explore a wide space of possible algorithm variants, with the goal to identify a set of~\textit{elite configurations} that perform best over a set of optimization problems. Then, the module importance is represented as the module's frequency in the set of elite configurations. This process is repeated by iteratively adding modules to the space~\textit{irace} explores. The outcomes emphasize that by integrating modules, the dynamics between the modules underwent changes, which became apparent through visualization of the results from their frequencies in the elite configurations. Nevertheless, a comprehensive quantitative analysis isolating all effects of the modules (i.e., including individual, pairwise, triple effects, etc.) is missing. 

Assessing the importance of various modules for the performance of modular optimization algorithms is analogous to evaluating the importance of hyperparameters in Machine Learning (ML) models. A seminal work in this domain is~\cite{hutter2014efficient}. Their approach is built on top of the classic technique of functional analysis of variance, previously introduced for high-dimensional functions of dependent variables~\cite{hooker2007generalized}. To quantify the influence of hyperparameters and their interactions in ML algorithms,~\cite{hutter2014efficient} expands the concept of the functional-ANOVA (f-ANOVA) framework as an efficient linear-time algorithm for calculating the interaction effects of hyperparameters. The efficacy of this approach was then used to investigate highly parametric ML frameworks and combinatorial solvers designed to optimize NP-hard problems. 

\section{Assessing module importance with f-ANOVA}
\label{sec:fANOVA}
f-ANOVA decomposes the observed variance $V$ of a response variable $f: \Theta_1 \times \cdots \times \Theta_n \to \mathbb{R}$ (in our case: algorithm performance) into additive components $V_{U}$, attributing each component to a corresponding subset of the inputs $U \subseteq \{1, \ldots, n\}$ (in our case: algorithm modules), and guarantees that the components add up the total variance of $f$ in the end: 
\begin{equation}
    V = \sum_{U \subseteq \{1, \ldots, n\}} V_U
\end{equation}
Fully decomposed, we can represent the variance in algorithm performance as: 

\begin{equation}
\label{eq:explained_variance}
\begin{aligned}
     V &=  V_1 + \cdots + V_n \\
    &\quad + V_{1,2} + \cdots + V_{1,n} + \cdots + V_{n-1,n} \\
    &\quad + \cdots \\
    &\quad + V_{1,\ldots,n}
\end{aligned}
\end{equation}

The components $V_{U}$ for $|U| = 1$ are called \textit{individual effects} and represent the variance in the algorithm performance caused by varying a single module's options. 
The components for $|U| > 1$ capture the \textit{interaction effects} between all modules in $U$. They represent the variance in the algorithm performance caused by varying the options for all the modules in $U$. 

To obtain this decomposition, we must be able to efficiently compute the effects over arbitrary sets of modules. In~\cite{hutter2014efficient} the authors provide an efficient and exact method for calculating the individual,  pairwise, and triple interaction effect of algorithm hyperparameters by performing variance decomposition using a random forest model~\cite{biau2016random}. In our case, the hyperparameters are analogous to the modules.

\section{Experimental design}
\label{sec:experiments}
In this section, we provide details on the experimental setup. We first describe our data including the algorithm portfolio, the problem portfolio, and the algorithm performance data. Next, the datasets used by the f-ANOVA approach for quantifying module efefcts are presented in more detail.

\subsection{Experimental data}  
We use publicly available data from the study presented in~\cite{kostovska2023using}. This repository contains performance data of 324 modCMA variants and 576 modDE variants, generated with the modular frameworks from~\cite{de2021tuning} and~\cite{DBLP:journals/corr/abs-2304-09524}, respectively. See Tables~\ref{tbl:modCMA_module_description} and~\ref{tbl:modCMA_modules} for a summary of the six modCMA modules, and Tables~\ref{tbl:modDE_module_description} and~\ref{tbl:modDE_modules} for the seven modDE modules considered in our study. Using the IOHexperimenter platform~\cite{iohexperimenter}, each variant is run 10 independent times on the first five instances of each of the 24 single-objective, noiseless black-box optimization problems from the BBOB benchmark suite~\cite{bbobfunctions} on the COCO platform~\cite{hansen2020coco}, in dimension $d\in\{5,30\}$. For each run on each problem instance, we calculate the~\textit{target precision} defined as the difference between the best-obtained solution and the global optimum. We consider the median target precision of the ten runs obtained after a budget of $100d$, $500d$, and $1500d$ function evaluations, respectively. Next, we transform the median target precision for each variant and each problem instance into logarithmic space, which is further used in our experiment referred to as the \textit{solution precision}. 

\begin{table*}[ht]
\centering
\caption{Short description of the role of modules in modCMA.}
\label{tbl:modCMA_module_description}
\begin{tabular}{ll}
    \hline
    Module & Description \\
    \hline
    1. base\_sampler &  The mechanism for sampling new candidate solutions around the current estimate of the optimum.\\
    2. elitism & The mechanism ensures that the best individual/s from the current population is preserved into the next generations.\\
    3. local\_restart & If the optimization process stagnates, a "restart" is performed from a different point in the search space.\\
    4. mirrored\_sampling & For every new sampled solution its mirrored image is added to the population. \\
    5. weights\_option & Recombination weights control how solutions from the current population are merged to create new individuals.\\
    6. step\_size\_adaptation & Strategy for adapting the algorithm's parameters which control the search process. \\
    \hline
\end{tabular}
\end{table*}

\begin{table}[ht]
\centering
\scriptsize % smaller text in table
{
\caption{Available options for the six modCMA modules evaluated in our experiments (324 combinations in total).}
\label{tbl:modCMA_modules}
\begin{tabular}{lc}
    \hline
    Module & Options \\
    \hline
    1. base\_sampler & Gaussian, Sobol, Halton \\
    2. elitist & off, on \\
    3. local\_restart & off, IPOP, BIPOP \\
    4. mirrored\_sampling & off, mirrored, mirrored pairwise \\
    5. weights\_option & default, equal, (1/2)\^{$\lambda$}  \\
    6. step\_size\_adaptation & csa, psr \\
    \hline
\end{tabular}
}
\end{table}

\begin{table}[ht]
\centering
\scriptsize % smaller text in table
{
\caption{Available options for the seven modDE modules evaluated in our experiments (576 combinations in total).}
\label{tbl:modDE_modules}
\begin{tabular}{lc}
    \hline
    Module & Options \\
    \hline
    1. adaptation\_method & off, shade, jDE \\
    2. crossover & bin, exp \\
    3. lpsr & off, on \\
    4. mutation\_base & rand, best, target \\
    5. mutation\_n\_comps & 1, 2 \\
    6. mutation\_reference & off, pbest, best, rand \\
    7. use\_archive & off, on\\
    \hline
\end{tabular}
}
\end{table}

\begin{table*}[ht]
\centering
\scriptsize % smaller text in table
{
\caption{Short description of the role of modules in modDE.}
\label{tbl:modDE_module_description}
\begin{tabular}{ll}
    \hline
    Module & Description \\
    \hline
    1. adaptation\_method & The strategy for adapting the algorithms' parameters which control the search process.\\
    2. crossover & Defines how the offspring solutions are generated from the parent solutions.\\
    3. lpsr & Strategy to gradually decrease the size of the population as the algorithm progresses. \\
    4. mutation\_base & This determines the base solution used in the mutation process, affecting the diversity of the population. \\
    5. mutation\_n\_comps & This parameter specifies the number of components (difference vectors) used in the mutation step.\\
    6. mutation\_reference & Determines which individuals are used as references in the mutation process.\\
    7. use\_archive & Controls whether past solutions are used in the evolution process.\\
    \hline
\end{tabular}
}
\end{table*}

\subsection{f-ANOVA datasets}
To quantify the individual and combined importance of modules in modular frameworks, we perform two distinct use-cases: one across the entire benchmark suite and another at the problem-specific level.

\textbf{Benchmark suite-level:} We compute the average solution precision for each variant across all 120 problem instances. This average solution precision serves as an estimation of the variant's performance across the entire BBOB benchmark suite. Subsequently, for modCMA, each variant is represented using six features (refer to the modules in Table~\ref{tbl:modCMA_modules}) and associated with the achieved mean solution precision. Likewise, for modDE, each variant is represented through seven features (refer to the modules in Table~\ref{tbl:modDE_modules}) and linked to the achieved mean solution precision across the complete benchmark suite. Following these transformations, we generate 12 datasets, two modular frameworks $\times$ two problem dimensions ($d=5,30$) $\times$ three different budgets of function evaluations ($100d, 500d,  1500d$), encompassing 324 variants for modCMA and 576 variants for modDE.

\textbf{Problem-level:} For this experiment we calculate the median solution precision of each variant, considering all five problem instances within each problem class. Following this, we generate distinct datasets for each problem by isolating the performance outcomes specific to that particular problem in two distinct problem dimensions ($d=5,30$). Consequently, we obtain 24$\times$2 datasets for each modular framework individually. Next, we repeat this for three different budgets (288 datasets in total). Similar to the benchmark suite-level approach, in the modCMA datasets, variants are characterized by six features associated with the module options, correlating with the median solution precision achieved on that specific problem. Conversely, in the case of modDE, seven features are utilized.

\subsection{f-ANOVA}
For variance decomposition analysis with a random forest model, we employed the implementation from~\cite{hutter2014efficient}. This method involves utilizing a random forest model as a regression tree predictor to forecast the performance of a variant using different module options. The complete dataset serves as the training data, upon which variance decomposition is directly applied using the trees within these forests. For a comprehensive understanding of why the training scenarios (such as cross-fold validation, or train-validation-test split) of the RF model are excluded, please refer to the original paper where detailed explanations are provided~\cite{hutter2014efficient}.

\section{Results and discussion}
\label{sec:results}
Here, the findings are categorized into two use cases: benchmark suite-level and problem-level. Within each use case, we display the outcomes of individual module effects, pairwise interaction effects, and triple interaction effects, for modCMA and modDE respectively, spanning two distinct problem dimensions and three different budgets.

\subsection{Benchmark suite-level}
\begin{table}[hb]
\centering
{
\scriptsize % smaller text in table
\caption{Cumulative fraction of variance (in \%) in the algorithm performance explained by the individual, pairwise, and triple interaction effects. 
}
\label{tab:effects_summary}
\begin{tabular}{lllrrrr}
\hline
algorithm & dim & budget & individual &  pairwise &  triple & total \\
\hline
modCMA & 5  & 100$d$   &  41.63 &     37.72 &   16.16 &  95.51 \\
      &    & 500$d$  &  42.05 &     38.89 &   15.94 &  96.88 \\
      &    & 1500$d$  &  43.88 &     34.87 &   16.34 &  95.09 \\
 & 30 & 100$d$  &  60.28 &     26.66 &   10.21 &  97.15 \\
      &    & 500$d$ &  51.16 &     30.40 &   13.70 &  95.26 \\
      &    & 1500$d$ &  54.31 &     28.95 &   12.43 &  95.69 \\
\hline      
modDE & 5  & 100$d$   &  80.88 &     13.12 &    4.45 &  98.45 \\
      &    & 500$d$  &  67.67 &     20.41 &    8.52 &  96.60 \\
      &    & 1500$d$  &  40.26 &     34.04 &   18.15 &  92.45 \\
 & 30 & 100$d$  &  68.76 &     19.26 &    8.13 &  96.15 \\
      &    & 500$d$ &  55.88 &     26.07 &   11.99 &  93.94 \\
      &    & 1500$d$ &  42.05 &     32.61 &   16.35 &  91.01 \\
\hline
\end{tabular}
}
\end{table}
Table~\ref{tab:effects_summary} displays the cumulative fraction of variance in the algorithm performance explained by the individual, pairwise, and triple module effects separately and provides information about how different types of module effects contribute to the total explained variance shown by the column ``total" in the table. The cumulative individual effect results from summing the individual effects of six modules for modCMA and seven modules for modDE, the cumulative pairwise effect is summed over $\binom{6}{2}=15$ module pairs of modCMA and $\binom{7}{2}=21$  of modDE, and finally, the cumulative triplet effect is calculated over $\binom{6}{3}=20$  module triples for modCMA and $\binom{7}{3}=35$ for modDE. The table contains the results for both algorithms modCMA and modDE, when considering two problem dimensions $d=5$ and $d=30$ and three different budgets 100$d$, 500$d$, and 1500$d$. The columns in the table are named correspondingly to the different effect types and the values denote \% of the total variance explained. 

Focusing on how the cumulative individual, pairwise, and triple effects vary across the different scenarios for modCMA, in $d=5$, the individual and pairwise effects are relatively balanced, while the triplet effect is smaller, for all budgets. In higher dimensionality ($d=30$) the individual effects tend to amplify.

Next, we will look at how the cumulative individual, pairwise, and triple effects vary for modDE. As the budget increases, the cumulative individual effect tends to decrease gradually, while the pairwise and triple effects increase correspondingly. This trend is consistent across both dimensions. Furthermore, comparing the same type of effects at different dimensions (5 vs. 30) for the same budget, there is a noticeable decrease in the individual effects and an increase in the pairwise and triple effects as the dimensionality increases. For the largest budget, the individual and pairwise effects even become balanced. 
Overall, for modDE, these observations indicate that the interactions between the modules (pairwise and triple effects) play a more significant role in the algorithm's performance as both the problem dimensionality and the budget for the algorithm increase.

Comparing modCMA and modDE, while both algorithms exhibit different shifts in the importance of individual, pairwise, and triple effects with changes in both dimensionality and budget, modDE experiences a more pronounced transition from individual to interaction effects. In contrast, modCMA maintains a stronger emphasis on individual effects, particularly in higher dimensions, even as the budget increases.

By summing all computed effects (cumulative individual, pairwise, and triplet) we get the total explained variance in the algorithm performance (presented in Table~\ref{tab:effects_summary}, the ``total" column). From this data, we can conclude that the total explained variance for both problem dimensions, all budgets, and the two investigated frameworks, is over 91\%, meaning that the results capture a comprehensive picture of how the modules and their interactions influence the algorithm performance. The rest of the variance in the algorithm performance is due to the complex interaction of more than three modules.

Next, we perform a more detailed analysis to understand the individual, pairwise, and triplet effects.

\begin{figure}[!htbp]
\centering
\begin{subfigure}[b]{0.24\textwidth}
\includegraphics[width=\textwidth]{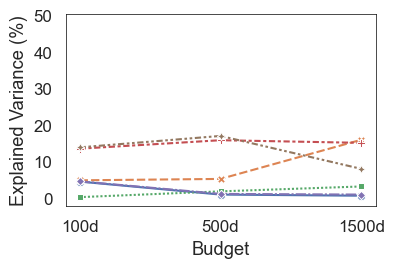}
\caption{dim=5, modCMA}\label{fig:CMA-ES_main_effect_5}
\end{subfigure}
\begin{subfigure}[b]{0.24\textwidth}
\includegraphics[width=\textwidth]{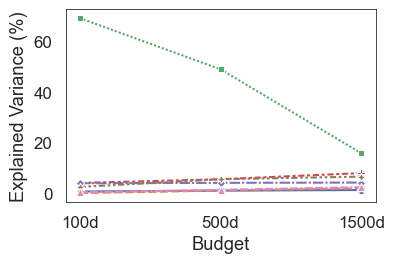}
\caption{dim=5, modDE}\label{fig:DE_main_effect_5}
\end{subfigure}
\begin{subfigure}[b]{0.24\textwidth}
\includegraphics[width=\textwidth]{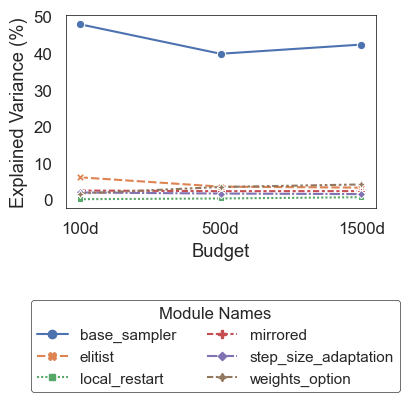}
\caption{dim=30, modCMA}\label{fig:CMA-ES_main_effect_30}
\end{subfigure}
\begin{subfigure}[b]{0.24\textwidth}
\includegraphics[width=\textwidth]{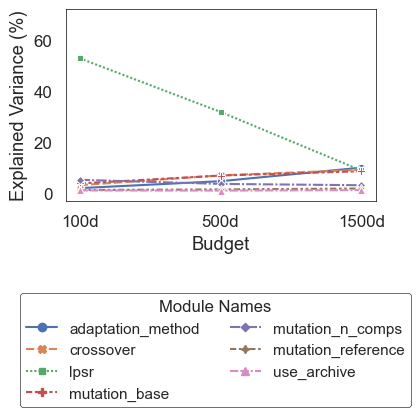}
\caption{dim=30, modDE}\label{fig:DE_main_effect_30}
\end{subfigure}
\caption{Variance in the algorithm performance explained by the individual effects of the algorithm modules of modCMA and modDE, in problem dimension $d=5$ and $d=30$ and for budgets of 100$d$, 500$d$, and 1500$d$ function evaluations.}
\label{fig:main_effects}
\end{figure}

\subsubsection{Individual module effects}
Figures~\ref{fig:CMA-ES_main_effect_5} and~\ref{fig:CMA-ES_main_effect_30} illustrate the portion of variance explained in the performance of modCMA by the individual effects of each of its modules, at two different problem dimensions (5 and 30) respectively, across three different budget (100$d$, 500$d$, 1500$d$). The same information for modDE is presented in Figures~\ref{fig:DE_main_effect_5} and~\ref{fig:DE_main_effect_30}. The $y$-axis presents the percentage of explained variance in algorithm performance, while the $x$-axis presents the budget used for evaluating the algorithm performance. The colored lines distinguish between the different modules. The legend is displayed below the plots, for each algorithm correspondingly. A summary of the key findings about how the effects are distributed among the modules is presented through the bullet points below:

\textbf{modCMA} - In the low dimensional case for the lower budgets we can observe that \textit{weights\_option} and \textit{mirrored} are the most important modules individually, explaining around 15\% of the variance in the algorithm performance each, where for the other the modules the explained variance varies between 1\% and 5\%. The relative importance of \textit{weights\_option} suggests that the recombination procedure which is impacted by this module is rather important early in the search. Upon examining the larger budget of 1500$d$ the individual importance of \textit{elitism} becomes more noticeable, as the algorithm has had time to converge, and potentially get stuck in local optima on the multi-modal functions. The importance of \textit{mirrored} remains relatively consistent, suggesting that its variance-reduction effects are beneficial throughout the search. In the higher dimensional case ($d$=30), irrespective of the budget allocated for evaluating the algorithm performance, consistent patterns emerge for the modules' importance. The \textit{base\_sampler} module accounts for a large fraction of the variance in algorithm performance on its own, while the remaining modules remain negligible. The reason for this is the instability of the Halton sampling procedure which when employed within this version of modCMA, causes biases in the sampling directions when the search dimensionality increases.
  
\textbf{modDE} - The primary module attributing to the variance in the algorithm performance is \textit{lpsr}. At the lower budgets, it explains a substantial part of the algorithm's performance variance (approximately 60\%). As the budget increases, this percentage diminishes to approximately 20\% for the largest budget. As noted in~\cite{DBLP:journals/corr/abs-2304-09524}, the reason for the large impact of \textit{lpsr} on modDE's performance is related to the default setting for population size, which is rather small in the wider context of DE. As such, enabling \textit{lpsr} changes the starting population size to be consistent with the popular L-SHADE variant~\cite{tanabe2013success} and leads to significant improvements in performance. For high dimensions, comparable trends are observed mirroring those in the low dimensionality scenario. 

\begin{figure*}[!htbp]
\centering
\begin{subfigure}[b]{0.45\textwidth}
\includegraphics[width=\textwidth]{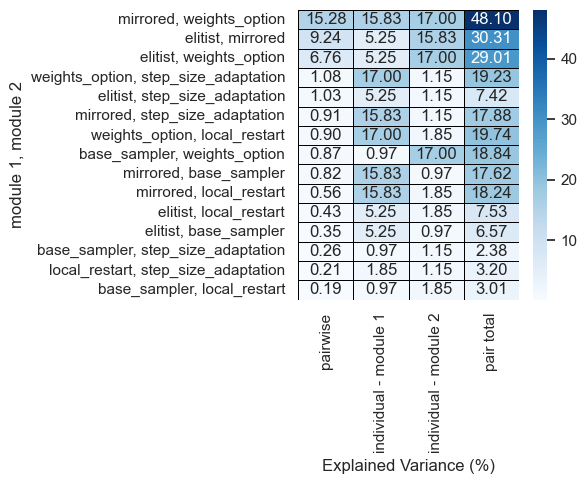}
\caption{dim=5, modCMA}\label{fig:CMA-ES_pairwise_effect_5}
\end{subfigure}
\begin{subfigure}[b]{0.45\textwidth}
\includegraphics[width=\textwidth]{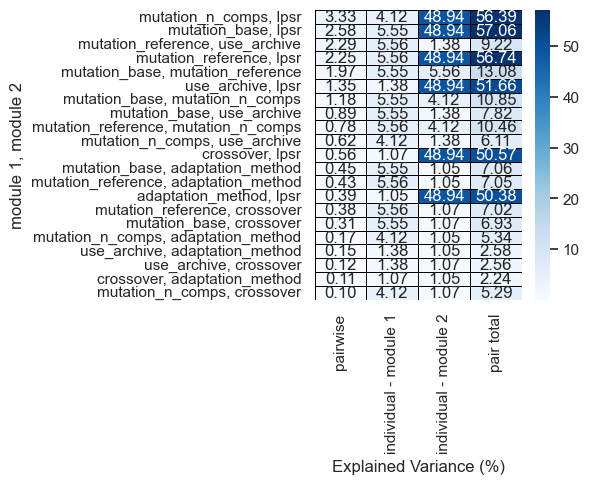}
\caption{dim=5, modDE}\label{fig:DE_pairwise_effect_5}
\end{subfigure}
\begin{subfigure}[b]{0.45\textwidth}
\includegraphics[width=\textwidth]{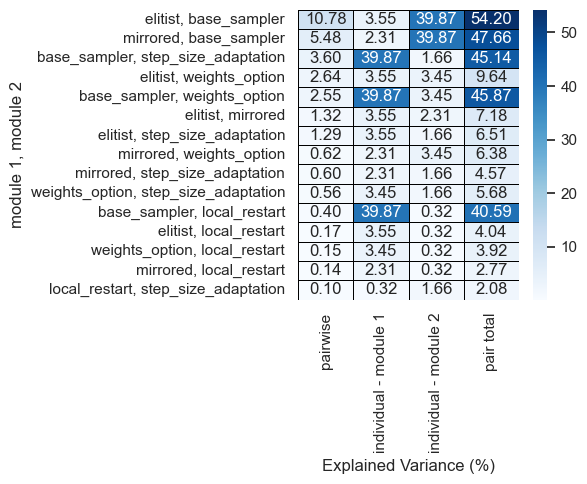}
\caption{dim=30, modCMA}\label{fig:CMA-ES_pairwise_effect_30}
\end{subfigure}
\begin{subfigure}[b]{0.45\textwidth}
\includegraphics[width=\textwidth]{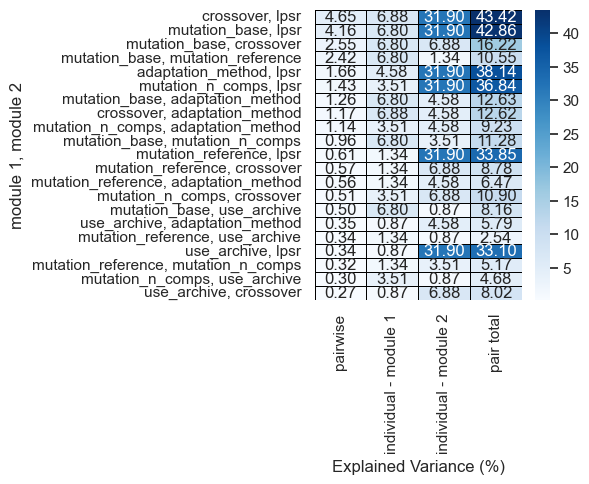}
\caption{dim=30, modDE}\label{fig:DE_pairwise_effect_30}
\end{subfigure}
\caption{Variance (\%) in the algorithm performance explained by the pairwise effects of the algorithm modules for a,c) modCMA and b,d) modDE in 5 and 30 dimensions respectively, for a budget of 500$d$.}
\label{fig:pairwise_effects}
\end{figure*}
\subsubsection{Pairwise module effects} Let us assume that we have a pair of modules ($\Theta_1, \Theta_2$). The total percentage of variance this pair explains can be calculated as $V_{pair\ total} = V_1 + V_2 + V_{1,2}$. The initial two terms denote the individual importance of the modules, previously examined in the analysis of individual module effects. The last term, the pairwise importance, offers insight into the portion of the explained variance stemming from their interaction effect. The results are presented for one budget (500$d$) due to the page limit, while for the other budgets, the results are available in our repository. 

Figures~\ref{fig:CMA-ES_pairwise_effect_5} and~\ref{fig:CMA-ES_pairwise_effect_30} depict the percentage of explained variance attributed to each module pair of modCMA for $d=5$ and $d=30$, respectively. The heatmap row headers display the module names involved in each pair, while the columns represent: the explained variance resulting from their interaction (the pairwise column), the individual module explained variances (individual - module\_1 and individual - module\_2 columns), and the total explained variance summing all three (the pair total column). The pairs are sorted in decreasing order based on the pairwise column. The results suggest an interaction between the \textit{mirrored} and \textit{weights\_option} module pair for $d=5$, contributing 15.3\% of the overall explained variance. The explained variance of the interaction is almost equal to each individual module's explained variance. Examining the rest of the interactions, we can note that involving the \textit{elitist} module, which by itself is not very important, with either \textit{mirrored} or \textit{weights\_option} in combination, their interaction contributes an additional 9.2\% and 6.8\% in the explained variance, respectively. All other pairwise interactions among the modules fall within similar ranges of explained variance, up to 1\%. 
For $d=30$, the most impactful pair interaction emerges from the \textit{elitist} and \textit{base\_sampler} modules, accounting for 10.8\% of the overall explained variance. Next, \textit{mirrored} and \textit{step\_size\_adaptation} explain a very small fraction of the variance on their own (their individual effect), while there is some additional value when combined with the \textit{base\_sampler}. It is noteworthy that pairs involving \textit{base\_sampler} yield the highest total explained variance, a result directly influenced by this module's notably high individual explained variance. Meanwhile, the remaining module pairs exhibit contributions ranging from 0\% to 3\% toward explaining the algorithm's performance variance.

The analyses for modDE are displayed in Figures~\ref{fig:DE_pairwise_effect_5} and~\ref{fig:DE_pairwise_effect_30} for $d=5$ and $d=30$, respectively. Across problem dimensions, it is evident that the variance explained by pairs of module interactions ranges between 0\% and 3.3\% for $d=5$ and between 0\% and 4.6\% for $d=30$. Notably, concerning the total explained variances for both problem dimensions, pairs involving \textit{lpsr} yield higher total explained variances due to this module's notably higher individual explained variance. While the module pairs without \textit{lpsr} explain relatively low amounts of variance overall, it is interesting to note that the combinations with the highest explanatory power are all modules that impact the mutation process. This suggests that picking the right combinations of mutation modules is important in DE, especially in low dimensions. In higher dimensions, the differences between binomial and exponential crossover become more significant, which is also reflected in the fact that combinations involving this module become relatively more impactful in the d=30 case.

\subsubsection{Triplet module effects}
Here the percentage of the variance in the algorithm performance that is explained by the interaction of three modules is presented. Let us assume that we have a triplet of modules ($\Theta_1, \Theta_2, \Theta_3$). The total percentage of variance a triplet explains can be calculated as $V_{total\ triplet} = V_1 + V_2 + V_3 + V_{1,2} + V_{1,3} + V_{2,3} + V _{1,2,3}$. The terms excluding the last one denote the portion of the explained variance stemming from the modules' individual and pairwise interactions, previously examined in the analysis of the individual and pairwise effects, while the last term is the portion of the explained variance stemming from the triplet interactions of the modules and is investigated next.  

Table~\ref{tab:triple_effects_total} displays the top five triplets, ranked according to the total percentage of explained variance associated with each module triplet. Additionally, it provides the contributions of the triplet interaction ($V_{1,2,3}$) in isolation. The results show that the highest-ranked triplet for modCMA, $d=5$, and 500$d$ budget consists of \textit{mirrored}, \textit{weights\_option}, and \textit{elitist}, which also is the triplet with the highest triplet effect. From the previous analysis, it is clear that \textit{mirrored} and \textit{weights\_option} have relatively large individual effects, meaning they are key drivers of the algorithm's success on their own. Further, their combination with the \textit{elitist} module which is not impactful on its own (refer to individual effects results), adds a lot of value to the total triplet explained variance, visible from the ``triplet" column and the pairwise interaction heatmaps in Figure~\ref{fig:CMA-ES_pairwise_effect_5}. For higher dimensions, the best triplet is highly influenced by the high individual importance of the \textit{base\_sampler} individual importance, while \textit{mirrored} and \textit{elitist} are not so impactful on their own retain their presence in the top triplet through interactions with \textit{base\_sampler}. 

In the case of modDE, the primary influence on the optimal triplet comes from one module, namely \textit{lpsr}. In lower dimensions, \textit{mutation\_base} and \textit{mutation\_reference} also play a contributing role through their interaction with \textit{lpsr}. In the higher dimensional case \textit{mutation\_base} and \textit{lpsr} remain in the best triplet, while \textit{crossover} replaces the \textit{mutation\_reference module}. 

\begin{table}[!ht]
{
\scriptsize % smaller text in table
\centering
\caption{Fraction of variance (in \%) in the algorithm performance explained by triple interaction effects of the algorithm modules. Only the five module combinations with the largest total importance are displayed.
}
\label{tab:triple_effects_total}
\begin{tabular}{lllrr}
\hline
 & $d$ & module1, module 2, module 3 &  triplet &  triplet \\
         &     &                             &          &  total \\
\hline
CMA & 5  & elitist, mirrored, weights\_option &                   8.19 &             77.54 \\
      &    & mirrored, weights\_option, step\_size\_adaptation &                   1.19 &             52.43 \\
      &    & mirrored, weights\_option, local\_restart &                   0.95 &             52.35 \\
      &    & mirrored, base\_sampler, weights\_option &                   1.16 &             51.93 \\
      &    & elitist, mirrored, step\_size\_adaptation &                   0.64 &             34.04 \\
  & 30 & elitist, mirrored, base\_sampler &                   2.96 &             66.27 \\
      &    & elitist, base\_sampler, weights\_option &                   1.71 &             64.55 \\
      &    & elitist, base\_sampler, step\_size\_adaptation &                   2.98 &             63.74 \\
      &    & elitist, base\_sampler, local\_restart &                   0.26 &             55.34 \\
      &    & mirrored, base\_sampler, weights\_option &                   1.03 &             55.31 \\
      \hline
DE & 5  & mutation\_base, mutation\_reference, lpsr &                   0.89 &             67.74 \\
      &    & mutation\_base, mutation\_n\_comps, lpsr &                   0.56 &             66.25 \\
      &    & mutation\_reference, mutation\_n\_comps, lpsr &                   0.34 &             65.31 \\
      &    & mutation\_reference, use\_archive, lpsr &                   0.79 &             62.55 \\
      &    & mutation\_base, use\_archive, lpsr &                   0.40 &             61.08 \\
 & 30 & mutation\_base, crossover, lpsr &                   1.18 &             58.10 \\
      &    & crossover, adaptation\_method, lpsr &                   0.63 &             51.46 \\
      &    & mutation\_base, adaptation\_method, lpsr &                   0.73 &             51.09 \\
      &    & mutation\_base, mutation\_n\_comps, lpsr &                   0.69 &             49.46 \\
      &    & mutation\_n\_comps, crossover, lpsr &                   0.34 &             49.22 \\
\hline
\end{tabular}
}
\end{table}

\subsection{Problem-level}

\begin{table*}[thp]
\centering
{
\scriptsize % smaller text in table
\caption{The combinations of modules with the largest total importance for the 5th, 15th, and 23rd BBOB problems in $d=5$ and $d=30$ for a budget of 500$d$. The results are presented separately for modCMA and modDE.}\label{tab:problem_fANOVA}
\begin{tabular}{llllll}
\hline
   &    &                                      modCMA &  &                                     modDE & \\
dim & f\_id & module 1, module 2, module 3 & triplet-total & module 1, module 2, module 3 & triplet-total\\
\hline
5  & 5  &  elitist, weights\_option, step\_size\_adaptation &            29.52 &  mutation\_base, mutation\_n\_comps, use\_archive &            23.73 \\
   & 15 &           elitist, base\_sampler, local\_restart &             62.4 &       mutation\_base, mutation\_reference, lpsr &            67.44 \\
   & 23 &              elitist, mirrored, weights\_option &            78.99 &         mutation\_reference, use\_archive, lpsr &            39.03 \\
30 & 5  &          elitist, base\_sampler, weights\_option &            81.08 &       mutation\_base, mutation\_reference, lpsr &            46.07 \\
   & 15 &                elitist, mirrored, base\_sampler &            69.93 &            crossover, adaptation\_method, lpsr &            57.71 \\
   & 23 &        elitist, mirrored, step\_size\_adaptation &            84.73 &                mutation\_base, crossover, lpsr &            57.59 \\
\hline
\end{tabular}
}
\end{table*}
The previous experiment yielded insights into the significance of modules and their interactions in explaining algorithm performance across the entire BBOB benchmark suite. Here, we are showcasing their importance at a problem-specific level. For this purpose, three problems from the BBOB benchmark suite are randomly selected: the 5th (linear slope), 15th (Rastrigin), and 23rd (Katsuuras). Table~\ref{tab:problem_fANOVA} illustrates the combinations of modules with the highest triplet-total effect on the performance, achieved on the 5th, 15th, and 23rd BBOB problems, across dimensions $d=5$ and $d=30$ and a budget of 500$d$. The outcomes are delineated individually for modCMA and modDE. For example, looking into the 5th problem in $d=30$, we can see that the interaction between the three modules \textit{elitis}, \textit{base\_sampler}, and \textit{step\_size\_adaptation} explains 81.0\% of the variance of the algorithm performance achieved on that problem. This variance is a sum of the individual effects (45.2\% (\textit{elitist} -M1), 13.7\% (\textit{base\_sampler} - M2), and 8.0\% (\textit{weights\_option)}  - M3), pairwise effects (5.6\% (M1, M3), 3.8\% (M1, M2), and 2.9\% (M2,M3)), and the triple effect (1.5\% (M1, M2, M3)). For the same problem in $d=5$, it appears that the optimal combination of modules elucidates roughly 30\% of the variance, underscoring the critical role of interactions among multiple modules (more than three) in achieving higher explained variances. Additionally, the combination of modules yielding the highest explained variance varies among different problems, which shows that different module interactions are important for solving different problems. We omit to present detailed results for each individual, pairwise, and triplet effects on each problem, however, they are publicly available on our GitHub repository.

\begin{figure}[!htbp]
\centering
\begin{subfigure}[b]{0.23\textwidth}
\includegraphics[width=\textwidth]{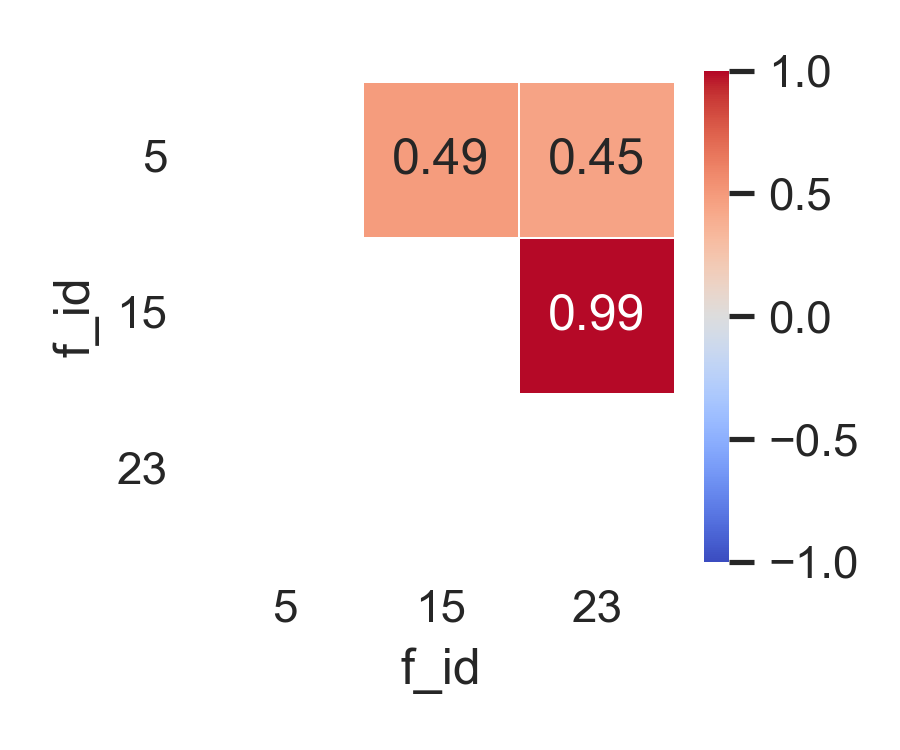}
\caption{dim=5, modCMA}\label{fig:CMA-ES_cosine_5}
\end{subfigure}
\begin{subfigure}[b]{0.23\textwidth}
\includegraphics[width=\textwidth]{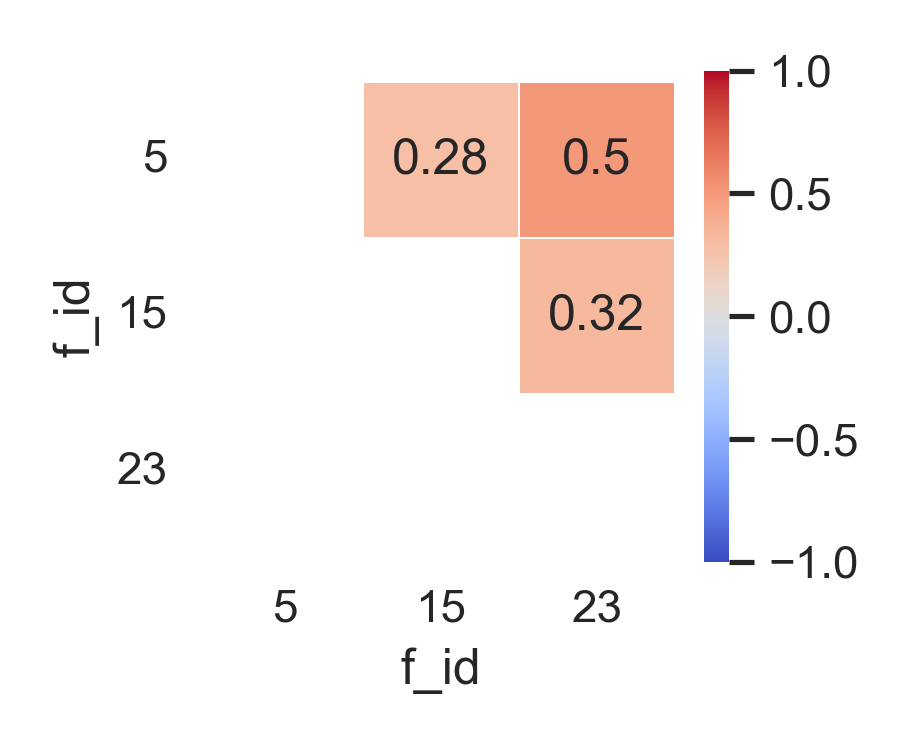}
\caption{dim=5, modDE}\label{fig:DE_cosine_5}
\end{subfigure}

\begin{subfigure}[b]{0.23\textwidth}
\includegraphics[width=\textwidth]{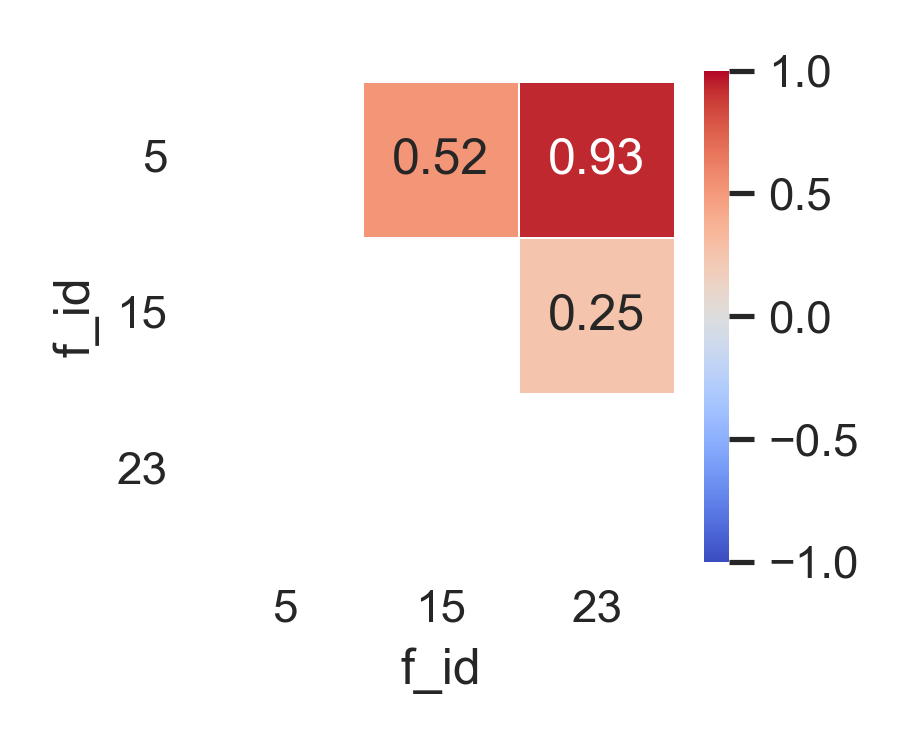}
\caption{dim=30, modCMA}\label{fig:CMA-ES_cosine_30}
\end{subfigure}
\begin{subfigure}[b]{0.23\textwidth}
\includegraphics[width=\textwidth]{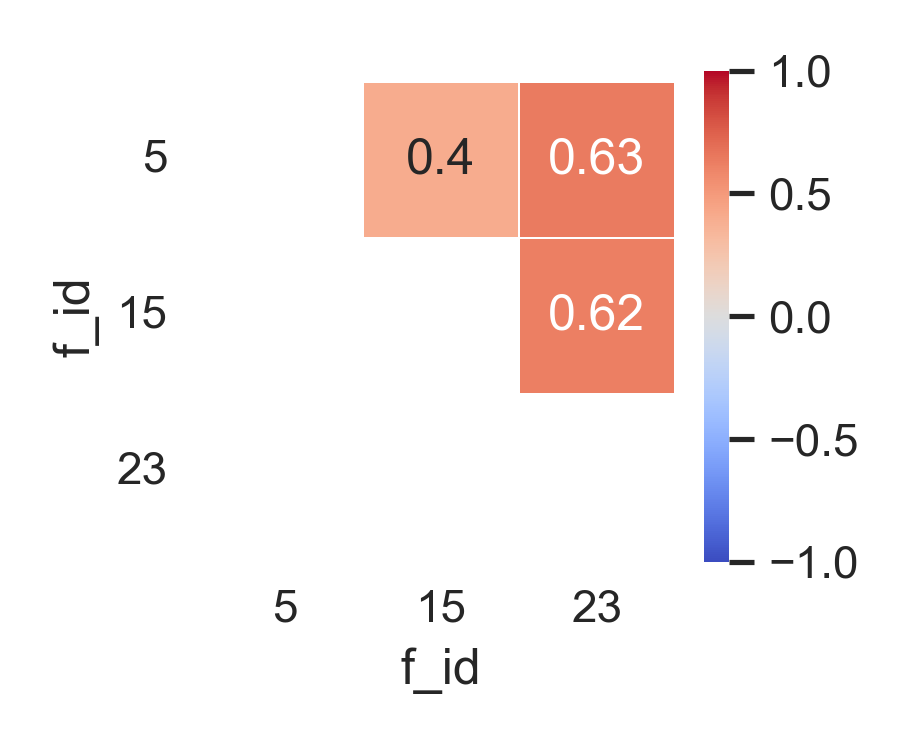}
\caption{dim=30, modDE}\label{fig:DE_cosine_30}
\end{subfigure}
\caption{Cosine similarity between three BBOB problems represented by the individual, pairwise, and triplet module effects for a,c) modCMA and b,d) modDE for a budget of 500$d$.}
\label{fig:cosine similarity}
\end{figure}

Given that we can compute individual (six for modCMA; seven for modDE), pairwise (15 for modCMA; 21 for modDE), and triplet effects (20 for modCMA; 35 for modDE) for each problem-level dataset, we can represent each problem using all quantified effects, 41 in the case of modCMA and 63 for modDE. By analyzing the similarity between problems based on these representations, we can explore which problems exhibit similar module interactions. To show this, we calculate the cosine similarity between the representations of the 5th, 15th, and 23rd problems in the case of modCMA separately for both dimensions (see Figures~\ref{fig:CMA-ES_cosine_5} and~\ref{fig:CMA-ES_cosine_30}) and for modDE (see Figures~\ref{fig:DE_cosine_5} and~\ref{fig:DE_cosine_30}). For modCMA in $d=5$, the module interactions show similarities ($\geq 0.9$) between the 15th and 23rd problems, whereas for $d=30$, similar module interactions are observed between the 5th and 23rd problems. For modDE, in $d=5, 30$, the module interactions provide different patterns for the analyzed problems, achieving similarity up to 0.6.

\section{Conclusions}
\label{sec:conclusion}
A wide range of modules is available for creating different variants of modular optimization algorithms designed for addressing single-objective black-box numerical optimization problems, but knowledge about the individual impact of each module on the algorithm's effectiveness, as well as how these modules interplay when integrated, is relatively insufficient. We employed the f-ANOVA framework~\cite{hutter2014efficient} to investigate the influence of modules and their interactions. Our focus was on two modular frameworks, namely modular Covariance Matrix Adaptation - Evolutionary Strategy (modCMA) and modular Differential Evolution (modDE), applied to 24 problems from the BBOB benchmark suite in dimensions 5 and 30. With 324 variants for modCMA and 576 for modDE, our goal was to assess individual and interaction module effects in two distinct scenarios: one evaluating the modules' impact on solving the entire benchmark suite (encompassing all problem instances), and the other examining their influence on a problem-specific level.

We have shown that for a problem dimension of 5 and modCMA, \textit{weights\_option} and \textit{mirrorred} play a substantial role in explaining most of the variance, albeit not significantly larger than other modules. Conversely, in the context of a problem dimension of 30, \textit{base\_sampler} exhibits the highest individual effect, contributing to half of the explained variance in algorithm performance. For modDE, \textit{lpsr} emerges as the primary factor explaining most of the variance for both problem dimensions. When comparing modCMA and modDE, both algorithms exhibit changes in the importance of individual modules and their interactions with variations in dimensionality and budget. Notably, modDE experiences a more noticeable transition from individual to interaction effects, whereas modCMA follows the opposite trend. The problem-level findings suggest a positive indication for identifying problems where module interactions contribute equally to algorithm performance.

Some limitations of the approach are that the calculation of module interactions requires an exponential amount of time, thus we are not able to calculate interactions beyond three modules in a reasonable amount of time.

A practical use case of the proposed approach can be the post-hoc analysis of algorithm configuration (AC) where the most important algorithm parameters are identified. Another example is the reduction of the module space as we can select the sub-space of the most influential algorithm modules and perform AC, which may lead to more promising results.

In future work, we plan to investigate another modular framework, PSO-X~\cite{camacho2021pso}. We also aim to identify problems exhibiting analogous modular interactions and establish connections between these findings and the inherent landscape properties of the problems under consideration together with the algorithm behavior (i.e., trajectories). This approach will offer additional assistance to designers of modular frameworks, providing them with deeper insights into the specific modules and the nature of interactions required for addressing certain landscape properties. We are also planning to test another approach for assessing module importance, the Pearson divergence ANOVA (PED-ANOVA) to calculate cross-form hyperparameter importance in arbitrary spaces~\cite{ijcai2023p488}.

\bibliographystyle{IEEEtran}
\bibliography{references}

\end{document}